\documentclass[letterpaper, 10 pt, conference]{ieeeconf}  

\IEEEoverridecommandlockouts                              

\usepackage{graphics} 
\usepackage{epsfig} 
\usepackage{amsmath} 
\usepackage{amssymb}  
\usepackage[hidelinks]{hyperref} 
\usepackage{algorithmic}
\usepackage{algorithm}
\usepackage{subfigure}
\usepackage{multirow}
\usepackage{xcolor}

\usepackage{amsthm}
\theoremstyle{definition}
\newtheorem{mydef}{Definition}
\newtheorem{myassum}{Assumption}

\title{\LARGE \bf
Distributed Control of Multi-Robot Systems in the Presence of\\ Deception and Denial of Service Attacks}

\author{Sangjun Lee and Byung-Cheol Min
\thanks{The authors are with the SMART Lab, Department of Computer and Information Technology, Purdue University, West Lafayette, IN 47907, USA
	{\tt\small lee1424@purdue.edu | minb@purdue.edu}}%
}

\begin{document}

\maketitle
\thispagestyle{empty}
\pagestyle{empty}

\begin{abstract} 
This research proposes a distributed switching control to secure multi-robot systems in the presence of cyberattacks. Two major types of cyberattack are considered: \textit{deception attack} and \textit{denial of service (DoS) attack}, which compromise the integrity and availability of resources, respectively. First, a residual-based attack detection scheme is introduced to identify the type of attacks. Then, a switching control is designed to neutralize the effect of the identified attacks, satisfying the performance guarantees required for state consensus among robots.
For the type of a deception attack, coordination-free consensus protocols are designed to tune the weights of each robot in a way that uncompromised robots gain more weight than compromised robots. 
For the type of a DoS attack, leader-follower protocols that reconfigure the communication topology are utilized to transform the compromised robots into sub-robots following the leaders. 
The performance of the proposed approach is evaluated on the Robotarium multi-robot testbed. A full demonstration with extensive cases is available at \href{https://youtu.be/eSj0XS2pdxI}{https://youtu.be/eSj0XS2pdxI}.

\end{abstract}


\section{Introduction}
Robotic systems are increasingly using open networks for operation, which poses new challenges for robotic systems. For example, a multi-robot application typically exchanges information between sensors, actuators, and controllers through cyberspace, allowing potential security breaches \cite{6899663}. In particular, the hierarchical nature of multi-robot systems that commonly operate through supervisory control, such as the Robot Operating System (ROS), in unprotected communication channels makes itself more vulnerable to cyberthreats. Vulnerability increases further when many robotic applications run on an open source framework that is fully accessible to unauthorized users. Several research studies \cite{2014Humphreys,8462886} demonstrate the risks of cyberattacks in different types of robotic systems, and these risks could lead to worst-case scenarios, such as security breaches and safety issues.
\begin{figure}[t]
\centering
\includegraphics[width=2.7 in]{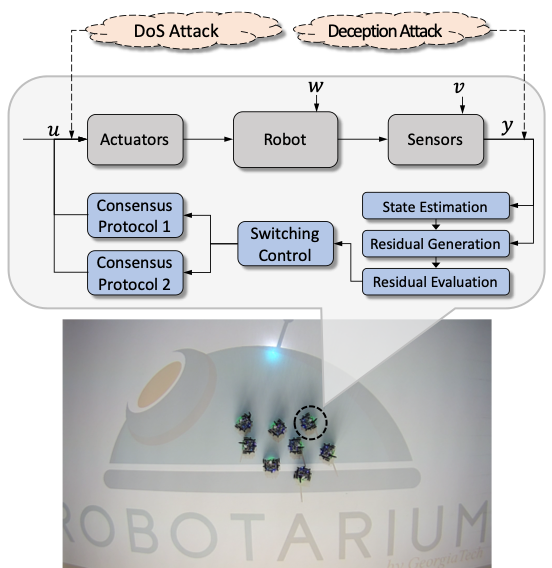}
\caption{Illustration of the proposed attack-resilient control scheme. (top) The block diagram of onboard control architecture for each robot. The blocks in blue are the main elements of attack detection and countermeasures. Solid lines represent information exchanges at the physical layer and dashed lines represent the cyber layer.
(bottom) A screenshot captured from experiments that depicts a team of homogeneous mobile robots reaching consensus at a common point.
}
\vspace{-2ex}
\label{fig:overview}
\end{figure}

The topic of multi-robot security against cyberattacks has received considerable attention in the past few decades \cite{4976621,8250942}, and there have been a number of attempts to provide solutions among a range of research communities.
In the context of the networked control systems, multi-robot systems are modeled as composed of a large number of simple systems interacting through communication channels \cite{6736104}. Recent studies seek solutions from various control design methods, such as robust control \cite{4577833} and adaptive control \cite{linan2018}, where cyberspace is represented as a discrete-time model and the underlying physical system is modeled after continuous-time dynamics. A common goal through these contributions is to achieve system resiliency and protect critical components from cyber threats.

Graph-theoretic methods have also been studied to address security problems in multi-robot systems \cite{6215022}. Generally, the main purpose is to formulate appropriate control policies to reach global consensus in the presence of attacks. Recent work in \cite{7867756} proposed a distributed control policy to achieve rendezvous with misbehaving agents. A distributed resilient consensus algorithm in \cite{6314937} showed that vehicles exponentially achieve the constrained consensus despite malicious attacks. A separate study \cite{7822915} proposed a dynamic strategy that ensures the communication network topology remains above a resilience threshold in the presence of cyberattacks.

However, there is still a lack of feasible solutions to implement in practice. 
For example, some prior studies \cite{8250942,lewis2013cooperative} rely on the assumption that each robot, including agents under attack, has unlimited communication range, allowing them to see all other robots continuously. This assumption requires high energy consumption or is unfeasible with a large number of robots and limited communication ranges. Thus, distributed control is a reliable alternative to ensure multi-robot security when depending only on local information of robots and their neighbors.
Another assumption commonly made is that the communication graph remains unchanged from the initial graph; however, the communication graph might be dynamic with time in many practical situations due to attacks \cite{li2014cooperative}.

The goal of cyberattack (or simply attack) is generally to disable the control system that connects various components associated with processes, causing security or safety issues. The proposed control scheme in this study is specifically designed to identify two major types of cyberattack: \textit{deception attack} and \textit{denial of service (DoS) attack} introduced in \cite{7011006,2009Amin}. The goal of \textit{deception attacks} is to compromise the integrity of control packets or measurements. \textit{DoS attacks}, instead, compromise the availability of resources. These two attack types are the main focus of this study since they cover most of possible attack scenarios in multi-robot systems including network jamming, false data injection, reply, and delay attacks. Specific strategies against each subtype are introduced in \cite{8936974,Bianchin2018,8431538,8691741}.

With the consideration of these limitations and design specifications, in this study, we propose a switching control to secure multi-robot systems in the presence of cyberattacks as illustrated in Fig. \ref{fig:overview}. The direction of arrival-aided detection schemes in \cite{lee2017smc,lee2018iros} are adapted and used to identify two types of cyberattacks. Then, a consensus-based switching control scheme is introduced to counter any unexpected deviation induced by attacks. For the first type of attacks, we utilize the coordination-free consensus protocols presented in \cite{ramviyas2018} to adaptively tune the weights on signal strength measurements based on the attack profile. This allows multi-robot systems to achieve global consensus by assigning more weight to the robots in normal operation than the compromised robots. A separate consensus protocol against the second type of attack is proposed using a leader-follower strategy that quickly reconfigures the communication topology to reassign the compromised robots as followers. This strategy ensures that there is at least one other neighbor in normal operation within sensor range, allowing each robot to switch its local control law between predefined consensus protocols in a fully distributed fashion.
In summary, this paper offers several distinct advantages over conventional approaches as follows:
\begin{itemize}
    \item We propose model-based attack detection techniques capable of identifying deception and DoS attacks.
    \item We propose countermeasures that satisfy the performance requirements in the presence of multiple attacks using a switching consensus control.
    \item We employ robots with limited sensor range, preventing a cascade failure.
    \item We validate the proposed methods via both experiments and simulations.
\end{itemize}

The remainder of this paper is organized as follows. 
First, models representing multi-robot systems' dynamic behaviors subject to attacks are presented in detail in Section \ref{sec:problem}.
Second, attack detection methodologies for each type are described and switching control schemes for a dynamic network topology with a limited sensor range are presented in Section \ref{sec:countermeasure}.
Third, the impact of the proposed approach is illustrated on a team of multiple mobile robots in Section \ref{sec:experiment}.
Finally, discussions and concluding remarks are presented in Sections \ref{sec:conclusion}.

\section{Problem Formulation}\label{sec:problem}
This section first provides a multi-robot consensus example in the presence of cyberattacks and then presents the interaction topology representing communication links among robots, followed by a description of multi-robot systems subject to attacks.

\subsection{Motivating Example}
\begin{figure*}[t]
	\centering
    \subfigure[final position]{\includegraphics[trim=0 -4cm 0 0, width=0.20\textwidth]{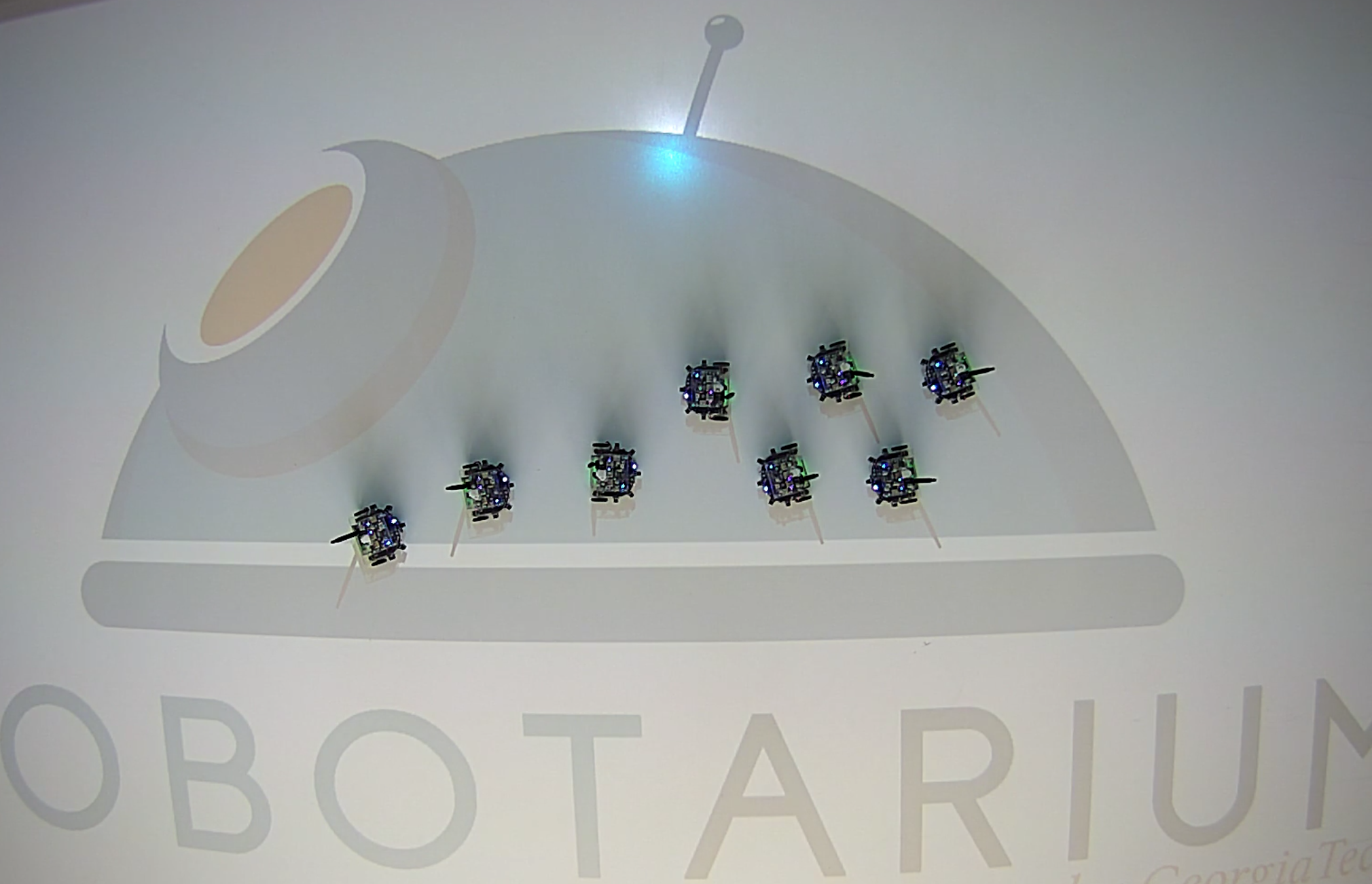}}
    \subfigure[trajectory]{\includegraphics[width=0.25\textwidth]{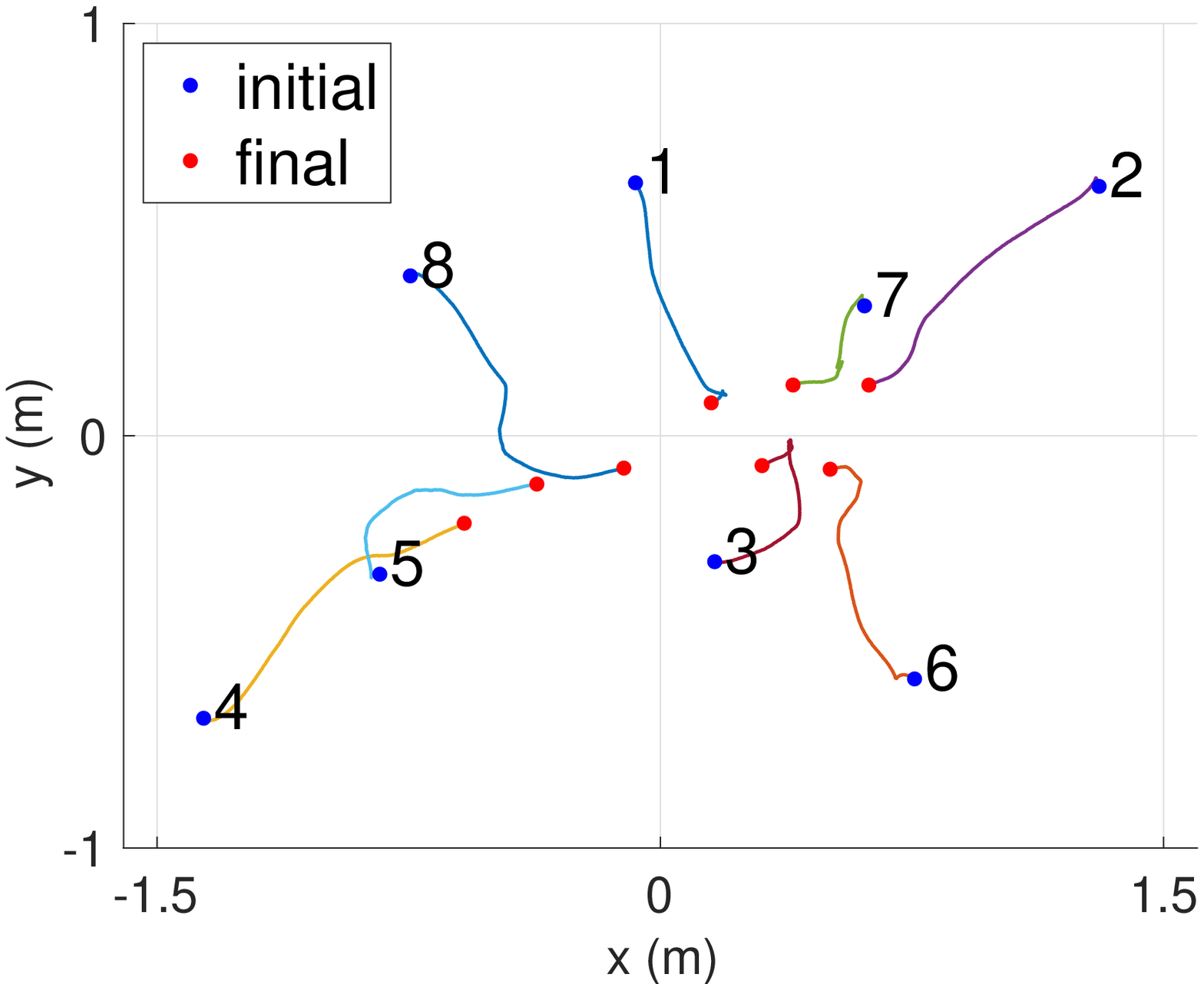}}
    \subfigure[final position]{\includegraphics[trim=0 -4cm 0 0, width=0.20\textwidth]{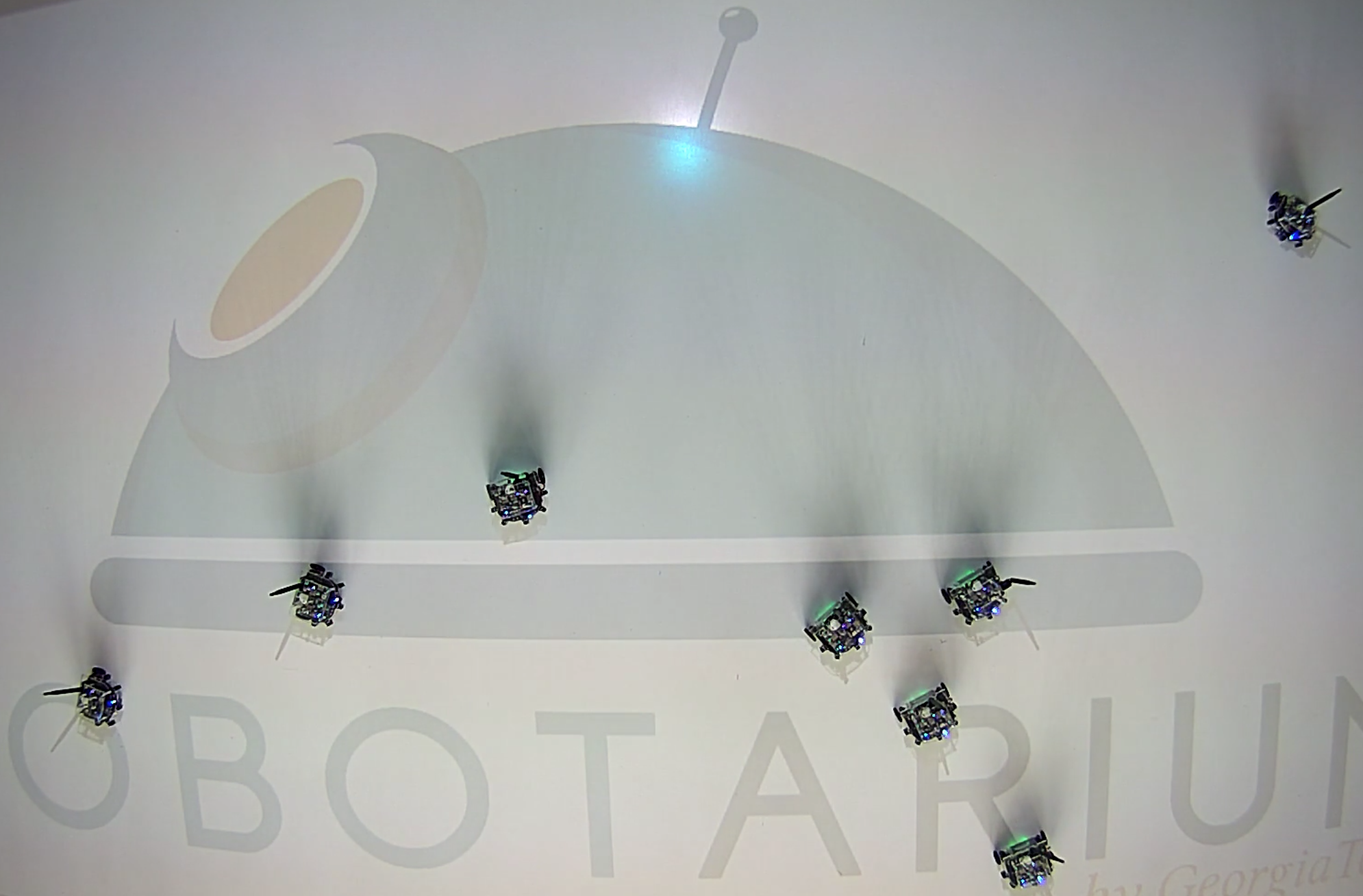}}
    \subfigure[trajectory]{\includegraphics[width=0.25\textwidth]{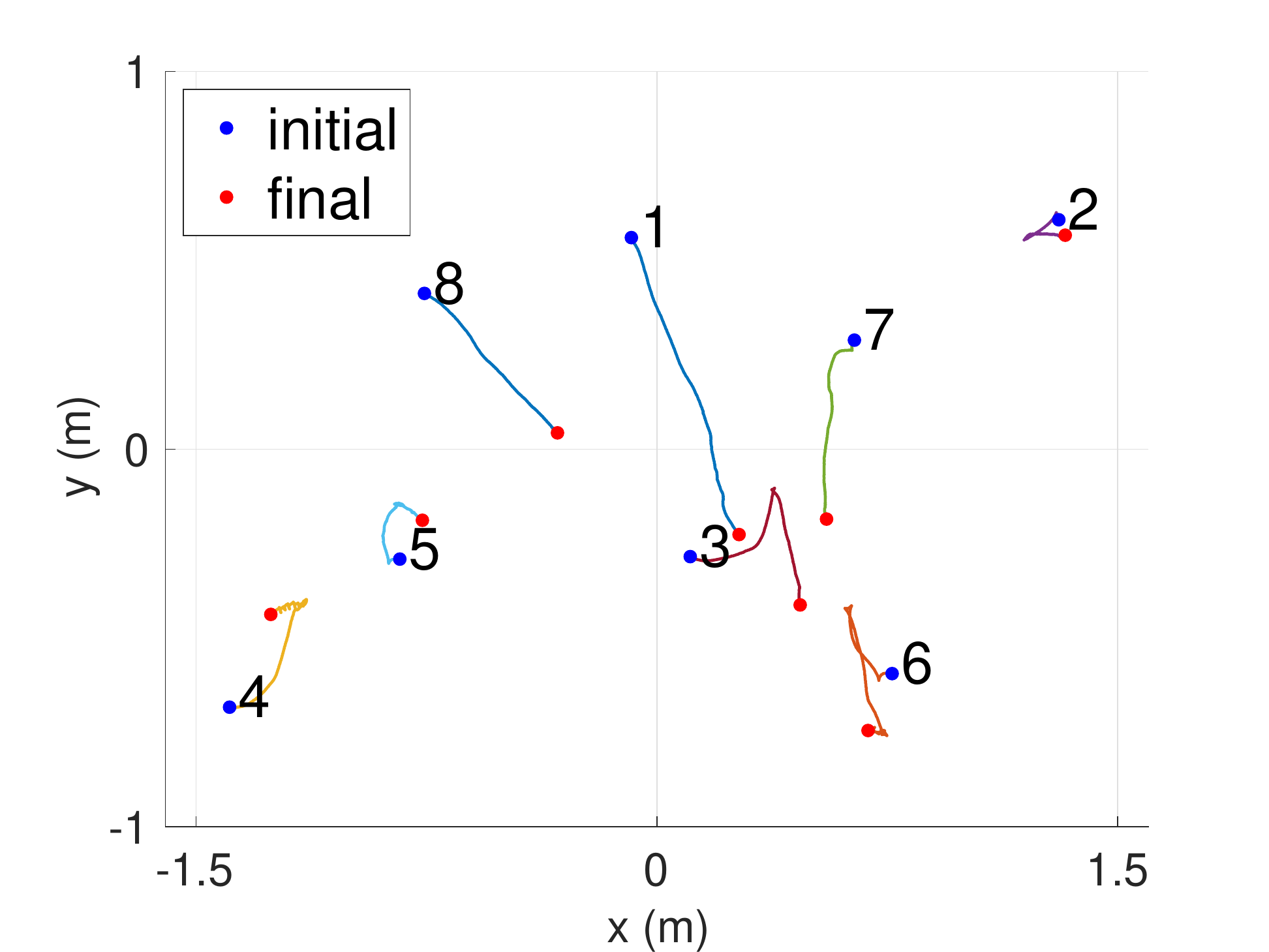}}
	\caption{Illustrative examples of multi-robot systems consensus problem. In (a) and (b), no compromised robots and consensus achieved. In (c) and (d), Robots $2$, $4$, $5$, and $8$ are compromised, resulting in the failure to reach state consensus. No proposed countermeasure was employed in this example.}
	\label{fig:C0}
\end{figure*}

A team of $8$ mobile robots initialized at a random pose and the aim is to  reach consensus at the center while maintaining a minimum safe distance between them. Each robot is able to access its own state and the local information from its neighbors within a sensor range. As shown in Fig. \ref{fig:C0} (a) and (b), the team achieved global consensus as each robot reached its goal position.

With the same condition, consider that there is a set of continuous unknown sensor reading measurement from an onboard sensor which is unable to identify. Such attack or unexpected measurement induces a significant deviation from the normal condition and results in the failure to reach state consensus as shown in Fig. \ref{fig:C0} (c) and (d). In real-life scenario (e.g. vehicle platooning), such attack can have fatal consequences. Thus, the main objective of this study is to design a switching control law that guarantees consensus in the presence of compromised robots in a team of $N$ mobile robots.

\subsection{Models}
Consider a group of $N$ identical robots. The interaction topology is represented as an undirected graph $\mathcal{G}=(\mathcal{V},\mathcal{E})$, where robots are represented as a nonempty finite node set $\mathcal{V}= \{ v_{1},v_{2},\ldots, v_{N} \}$ and the information exchanges between the robots are represented as edges $\mathcal{E} \subseteq \mathcal{V} \times \mathcal{V}$. The $i$-th robot has a set of neighbors that is the set of all vertices connected to node $i$, denoted as $N_{i}$ for $i=1,2, \ldots, N$. The weight associated with the edge $(v_{i},v_{j})$ in the edge set $\mathcal{E}$ is defined such that $a_{ij} \in \mathbb{R}$, $a_{ij} = 1$ if $(v_{i},v_{j}) \in \mathcal{E}$ and $a_{ij} = 0$ otherwise.
The model of the $i$-th robot is represented by a discrete-time LTI system with given matrices $A, B$, and $C$:
	\begin{equation}\label{eq:normal_sys}
	x_{k+1} = Ax_{k} + Bu_{k} + w_{k},
	\end{equation}
where $x_{k} \in \mathbb{R}^{n}$, $u_{k} \in \mathbb{R}^{p}$, and $w_{k} \in \mathbb{R}^{n}$ represent the state vector, a control input, and the process noise at time $k$, respectively. It is assumed that the process noise is an independent Gaussian random variable with zero mean and covariance $Q$ denoted as $w_{k} \sim \mathcal{N}(0,Q)$.
The model of sensors that measure the states of the robot in (\ref{eq:normal_sys}) is represented as:
    \begin{equation}\label{eq:normal_mea}
    y_{k} = Cx_{k} + v_{k},
    \end{equation}
where $y_{k} \in \mathbb{R}^{m}$ is the sensor measurement vector and $v_{k}$ is the measurement noise with zero mean and covariance $R$ such that $v_{k} \sim \mathcal{N}(0,R)$.

If an unauthorized user corrupts a random subset of outputs or introduces a delay to the communication link as shown in Fig. \ref{fig:overview}, the attack causes deviations at the physical layer such as $\tilde{y}\neq y$ or $\tilde{u}\neq u$, resulting in alterations to (\ref{eq:normal_mea}) which can be represented as follows:
    \begin{equation}\label{eq:ss_attack}
    y_{k}^{\alpha} = (1-p_{k})y_{k} + p_{k}y_{k-1} + \alpha_{k},
    \end{equation}
where $y_{k}^{\alpha} \in \mathbb{R}^{m}$ is the measured output corrupted by attacks, $p_{k} \in \mathbb{R}^{m}$ represents time delay attacks described as a Bernoulli distribution $p_{k}\sim\mathcal{B}(0,1)$, and $\alpha_{k} \in \mathbb{R}^{m}$ denotes additive attacks.
This insight leads us to formulate an assumption and definitions as follows.
\begin{myassum}\label{assumption1}
A set of unknown time-invariant attacks $ \mathcal{S}^{\alpha}  \subset \{ 1, 2, \ldots, N \}$, such that the corresponding attack signals $ \alpha_{i}(k) \in \mathbb{R}^{n}$ for arbitrary $i \in \mathcal{S}^{\alpha}$ at any $k$. On the other hand, the set of attack-free sensors $\mathcal{S}:=\{1,2,\ldots,N\} \setminus \mathcal{S}^{\alpha}$ holds that $\alpha_{i}(k) \equiv 0$ for $i \in \mathcal{S}$ at any $k$.
\end{myassum}
\begin{mydef}
The $i$-th robot $x_{i}\in \mathcal{V}$ is under a \textit{deception attack} if $\alpha_{k} \neq 0$ for $k \geq k_{\alpha}$ and $p_{k}=0$ for all $k$ in (\ref{eq:ss_attack}). This scenario is classified as \textit{Type 1}.
\end{mydef}
\begin{mydef}
The $i$-th robot $x_{i}\in \mathcal{V}$ is under a \textit{DoS attack} if $\alpha_{k}=0$ for all $k$ and $p_{k} \neq 0 $ for $k \geq k_{\alpha}$ in (\ref{eq:ss_attack}). This scenario is classified as \textit{Type 2}.
\end{mydef}
\begin{mydef}
A team of $N$ robots is said to achieve attack-resilient consensus if $||x_{k,i}-x_{k,j}|| =  0$ as $k \rightarrow \infty$ for all $ i,j=1,2,\ldots,N$ in the presence of attacks.
\end{mydef}

\subsection{Kalman Filter-based Consensus Control}
Suppose that the system in (\ref{eq:normal_sys}) and (\ref{eq:normal_mea}) satisfies the steady-state condition in a pre-attack state. Then a steady state Kalman filter provides an output prediction $y_{k}$ and $\hat{y}_{k}$, allowing the system to detect any significant deviation between pre-attack and post-attack states. The state estimate is given as:
	\begin{equation*}
	\hat{x}_{k+1} = A \hat{x}_{k} + B u_{k} + L(y_{k} - \hat{y}_{k}),
	\end{equation*}
where $L = PC^{T}(CPC^{T} + R)^{-1}$ with the covariance matrix denoted by $P = A[P-PC^{T}(CPC^{T}+R)^{-1}CP]A^{T} + Q$, and $(A,C)$ is detectable. The residual dynamics for the $i$-th robot is given as:
	\begin{equation*}
    r_{k} = y_{k}^{\alpha} - \hat{y}_{k}.
    \end{equation*}

Represent the output estimate as $\hat{y}_{k} = C \hat{x}_{k}$ and the estimate error as $e_{k} = x_{k} - \hat{x}_{k}$. Then, the output prediction error is defined as:
	\begin{equation}\label{eq:resi}
	r_{k+1} =(1-p_{k+1}) Ce_{k+1} + p_{k+1}y_{k} + \alpha_{k+1},
	\end{equation}
where the estimation error dynamics are given by $e_{k+1} = (A-LC) e_{k}$. 
This can be used to obtain the new information in $y_{k}$, which was not available in $y_{1}, \ldots, y_{k-1}$. 

Each robot described by (\ref{eq:normal_sys}) and (\ref{eq:normal_mea}) is able to access its own state and the local information from its neighbors within a sensor range. A consensus protocol for $i$-th robot is derived:
    \begin{equation}\label{eq:consensus}
    \dot{x}_{i} = (A+LC)x_{i} + \sum_{j=1}^{N} a_{ij}[BK(x_{i}-x_{j}) + L(y_{i}-y_{j})],
    \end{equation}
where $K$ is the feedback gain. Each robot updates its reference state at each instant $k$ as a weighted combination of its own current state and other measured states received from its neighbors denoted as $a_{ij}$ that is the $(i,j)$-th entry of the adjacency matrix. By defining $e_{i} = x_{i} - \sum_{j=1}^{N}a_{ij}(x_{i}-x_{j})$, the following dynamics can be derived:
    \begin{equation*}
    \dot{e}_{i} = (A+LC)e_{i}.
    \end{equation*}
Therefore, the consensus protocol solves the consensus problem for the robot described by (\ref{eq:normal_sys}) if the matrix $A+LC$ is a stability matrix. This consensus protocol underlies each agent until the detection mechanism identifies the type of attacks.

\section{Countermeasures Against Attacks}\label{sec:countermeasure}
In this section, distributed consensus control policies are designed to achieve attack-resilient consensus. For each type, residual-based attack detection schemes and state feedback consensus protocols are presented.

\subsection{Type 1. Deception Attacks}
If an attack fits the description of a deception attack in Definition 1, the output prediction error in (\ref{eq:resi}) where $p_{k}:=0$ becomes:
	\begin{equation*}
	r_{k+1} =Ce_{k+1} + \alpha_{k+1},
	\end{equation*}
and this detection problem is solved by tracking the historical behavior of the residual as shown in \cite{lee2018iros} as:
\begin{align}\label{eq:cusum}
	S_{k+1}=
	\begin{cases}
	\max{(0,S_{k} + |r_{k+1}|)} \text{ if } S_{k} \le \tau_{k} \\
	0 \text{ and } k_{\alpha}=k  \text{ if } S_{k} > \tau_{k},
	\end{cases}
\end{align}
where the decision function generates a global attack alarm time $k_{\alpha}$ at which test statistics $S$ exceeds a given threshold.

At the time step $k=k_{\alpha}$, when the detection scheme identifies any deception attack, a weighted bearing controller introduced in \cite{ramviyas2018} is applied to counter the attacks as:
\begin{equation*}
    u_{i} = \frac{1}{N} \sum_{j=1}^{N} w_{ij}(x_{i}-x_{j}) \text{ and } w_{ij} = \frac{1}{\gamma_{ij}-\gamma_{\tau}},
\end{equation*}
where $\gamma_{ij}$ denotes the received signal strength transmitted from robot $j$ measured at robot $i$, and $\gamma_{\tau}$ is the threshold of received signal strength power. The core idea of the control policy proposed for this case is to assign the compromised robots with less weight than the others. This strategy allows the robots in normal operation to attract the compromised robots toward them, ultimately leading to consensus. 
From the proof of consensus convergence in \cite{ramviyas2018}, it is easy to see that the $N$ robots in (\ref{eq:normal_sys}) and (\ref{eq:normal_mea}) reaches consensus under the protocol (\ref{eq:consensus}).

\subsection{Type 2. Denial of Service (DoS) Attacks}
In the case of a DoS attack like that in Definition 2, the output prediction error in (\ref{eq:resi}) where $\alpha_{k}:=0$ becomes:
	\begin{equation*}
	r_{k+1} =(1-p_{k+1}) Ce_{k+1} + p_{k+1}y_{k}.
	\end{equation*}
	
Consider that a stochastic process $z(k)$, mutually independent of $p_{k+1}$, is also a Bernoulli distributed white sequence with expected value $\mu$. During the pre-attack state, $\mu$ is equal to $\mu_{0}$. It would change to $\mu = \mu_{1} \ne \mu_{0}$ otherwise. Then, the detection problem needs to distinguish between two hypotheses:
	\begin{align*}
	\begin{split}
	\mathcal{H}_{0}: & \quad  z_{k} \sim \mathcal{B}(\mu_{0}, 1) \text{ for } k=1,...,k \\
	\mathcal{H}_{1}: & 
		\begin{cases}
		z_{k}  \sim \mathcal{B}(\mu_{0}, 1) \text{ for } k=1,...,k_{\alpha}-1 \\
		z_{k} \sim \mathcal{B}(\mu_{1}, 1) \text{ for } k \ge k_{\alpha},\\ 
		\end{cases}
	\end{split}
	\end{align*}
and a decision function for this case is similar to (\ref{eq:cusum}), which is omitted here for brevity.

A new controller is required because the controller designed for \textit{Type 1} attack is unable to handle measurement delay attacks. A strategy that enables the agents’ states to converge onto a reference trajectory is proposed, and it can be solved by leader-follower consensus approach. The detection scheme identifies the index of the compromised robots and transforms them into followers, while the robots in normal operation are assigned as leaders. 
The consensus protocol for the leaders follows (\ref{eq:consensus}), and the following distributed controller is proposed for each follower as:
    \begin{align}\label{eq:follower}
    \begin{split}
    \dot{x}_{i} &= (A+BF)x_{i} + cL\sum_{j=1}^{N} a_{ij}[C(x_{i}-x_{j}) - (y_{i}-y_{j})],\\
    u_{i} &= Fx_{i},
    \end{split}
    \end{align}
where $c>0$ is the coupling gain, and $L$ and $F$ are feedback gain matrices. The consensus problem can be solved by the controller (\ref{eq:follower}) with the feedback gain $F$ such that $A+BF$ is Hurwitz and $L=-Q^{-1}C^{T}$ where $Q>0$ is a positive-definite solution of the following linear matrix inequality:
\begin{equation}
    A^{T}Q+QA-2C^{T}C<0.
\end{equation}

Let $z_{f}:=[z_{1}^{T}, \cdots,z_{M}^{T} ]^{T}$, $z_{l}:=[z_{M+1}^{T}, \cdots,z_{N}^{T} ]^{T}$, and represent the Laplacian matrix associated with $\mathcal{G}$ as:
    \begin{equation*}
        \mathcal{L}=
        \begin{bmatrix}
        \mathcal{L}_{1} & \mathcal{L}_{2} \\
        0 & 0
        \end{bmatrix}.
    \end{equation*}
Under the assumption that each follower has at least one leader with a directed path, similar to \cite{meng2010distributed}, all the eigenvalues of $\mathcal{L}_{1}$ have positive real parts, each entry of $-\mathcal{L}_{1}^{-1} \mathcal{L}_{2}$ is nonnegative, and each row of $-\mathcal{L}_{1}^{-1} \mathcal{L}_{2}$ has a sum equal to one. The robots in (\ref{eq:normal_sys}) and (\ref{eq:normal_mea}) under the controller (\ref{eq:follower}) achieve state consensus if the following error asymptotically converges to zero:
\begin{equation}
    \epsilon = (\mathcal{L}_{1}\otimes I_{2n})z_{f}+(\mathcal{L}_{2}\otimes I_{2n})z_{l}.
\end{equation}
The dynamics of $\epsilon$ are given as:
\begin{equation}\label{eq:epsilon}
    \dot{\epsilon} = (I_{M} \otimes \mathcal{W}_{1}+c\mathcal{L}_{1} \otimes \mathcal{W}_{2})\epsilon,
\end{equation}
where 
    \begin{equation*}
        \mathcal{W}_{1}=
        \begin{bmatrix}
        A & BF \\
        0 & A+BF
        \end{bmatrix} \text{ and }
        \mathcal{W}_{2}=
        \begin{bmatrix}
        0 & 0 \\
        -LC & LC
        \end{bmatrix}.
    \end{equation*}
This follows that there exists a $Q>0$ satisfying:
\begin{equation*}
    Q\mathcal{H}+\mathcal{H}^{H}Q \le QA + A^{T}Q-2C^{T}C < 0,
\end{equation*}
where $\mathcal{H}=A+c\lambda_{i}LC$, which is Hurwitz. Therefore, the error dynamics in (\ref{eq:epsilon}) is asymptotically stable implying that state consensus is achieved.

\section{Validations and Results}\label{sec:experiment}
In this section, the proposed switching architecture is implemented on multi-robot consensus control problems with three different attack scenarios, and the results are provided. We consider that an intruder is able to synthesize each type of attacks, satisfying Assumption \ref{assumption1} and inject into a team of homogeneous robots through cyberspace.
\begin{figure*}[t]
	\centering
    \subfigure[]{\includegraphics[width=0.29\textwidth]{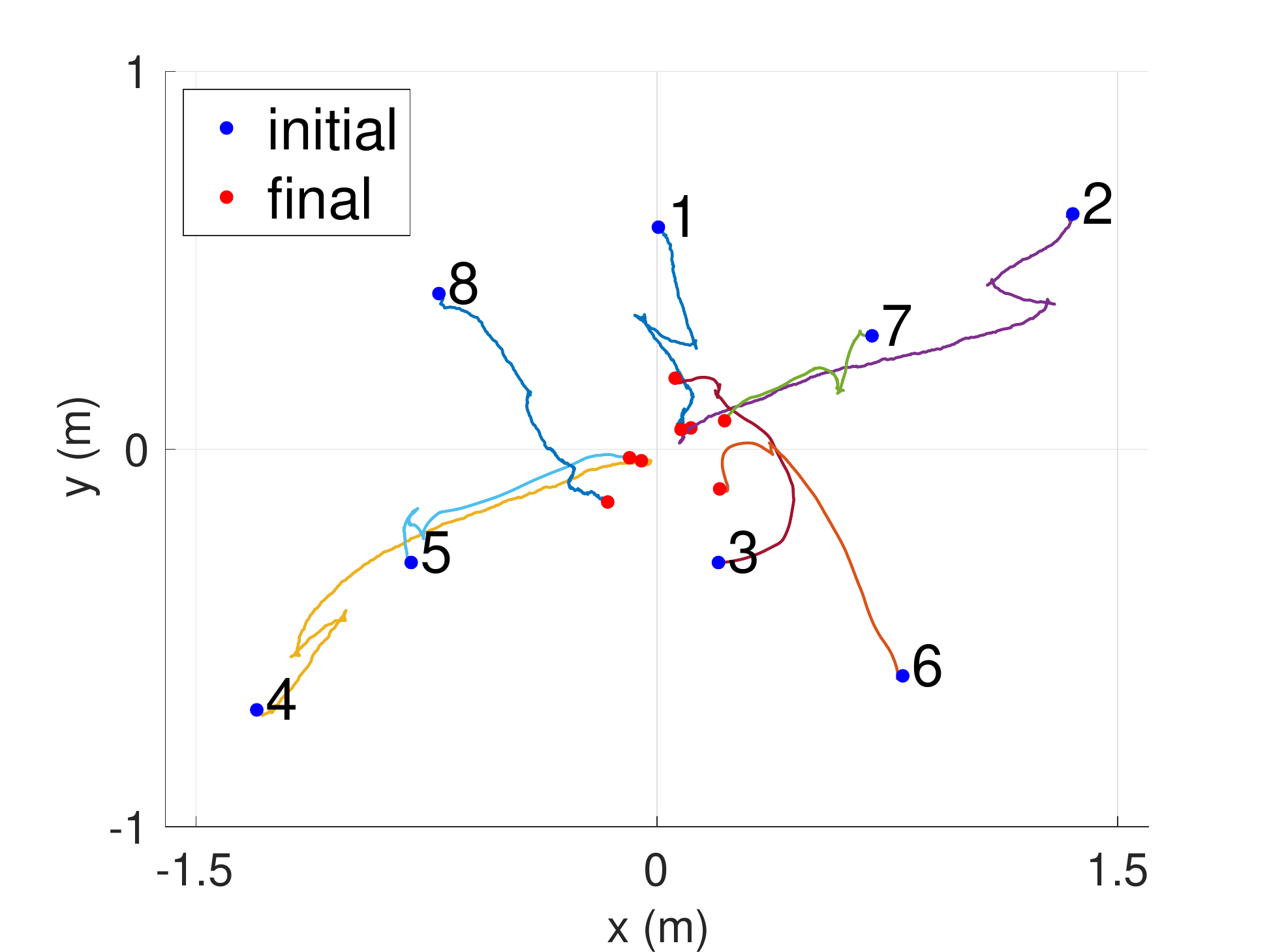}}
    \subfigure[]{\includegraphics[width=0.29\textwidth]{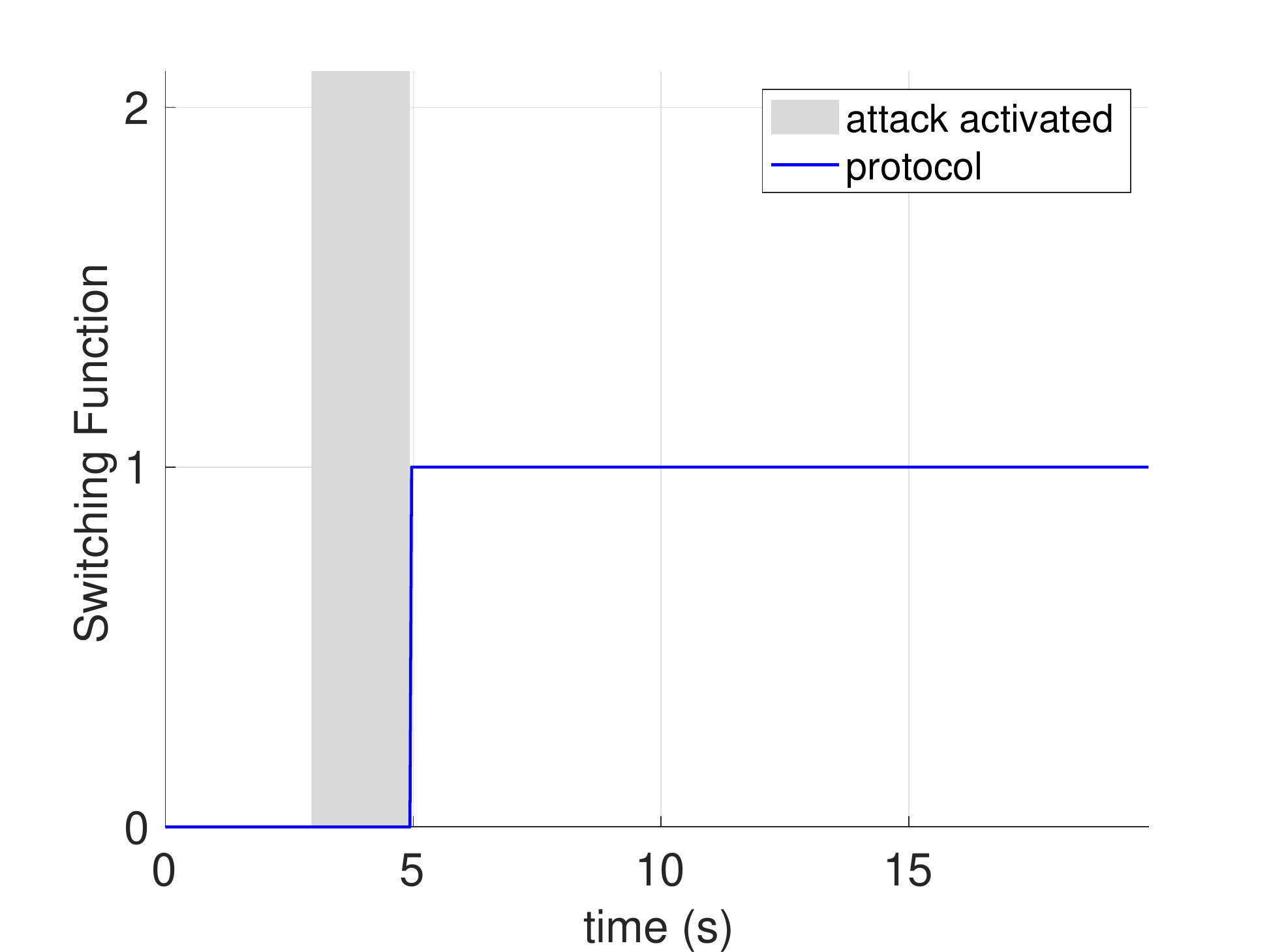}}
    \subfigure[]{\includegraphics[width=0.29\textwidth]{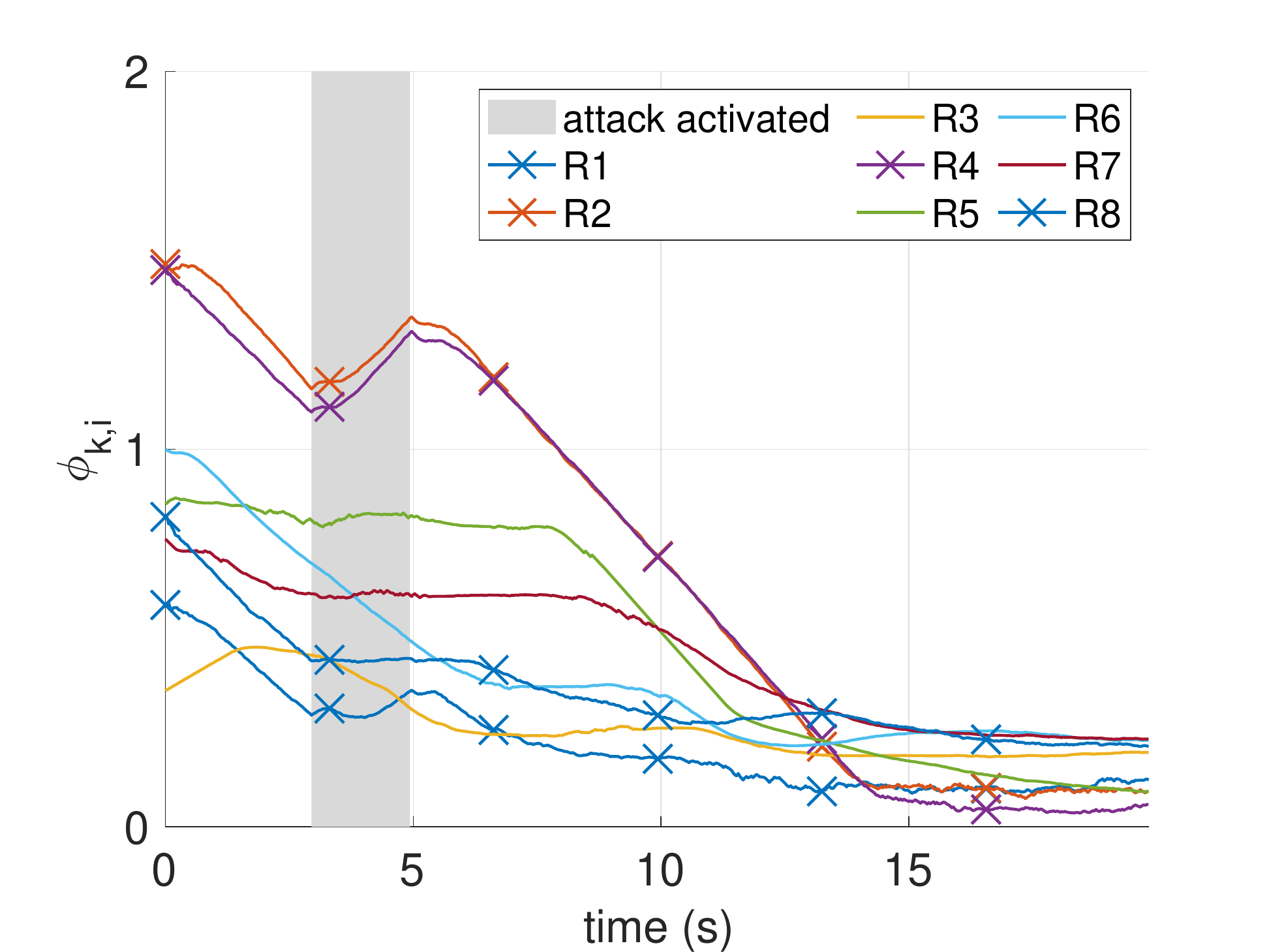}}
    \vspace{-1ex}
	\caption{Multi-robot consensus control in the presence of \textit{deception attack} (Robots $1, 2, 4$, and $8$ compromised).}
	\label{fig:C1}
\end{figure*}
\begin{figure*}[t]
	\centering
    \subfigure[]{\includegraphics[width=0.29\textwidth]{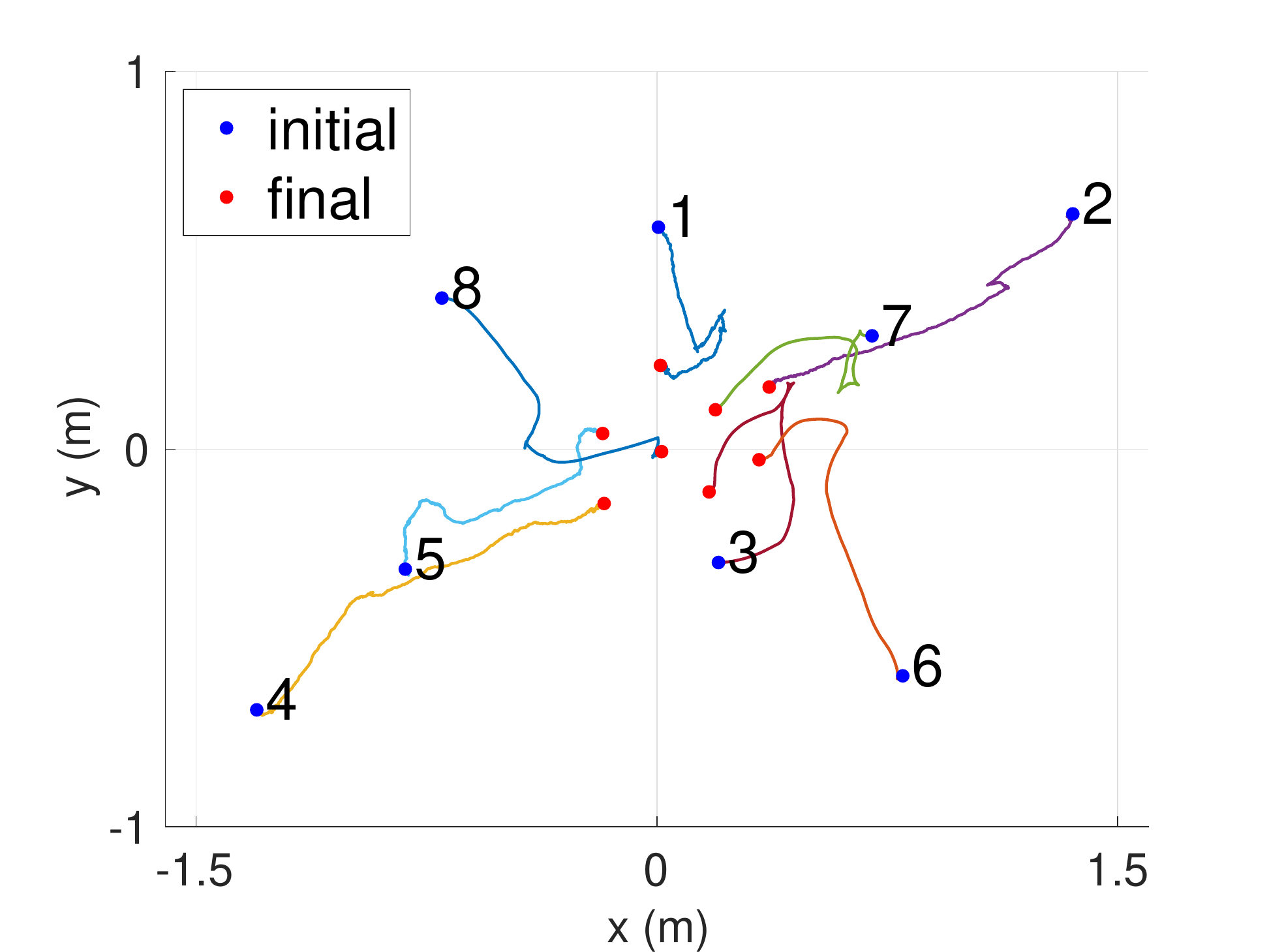}}
    \subfigure[]{\includegraphics[width=0.29\textwidth]{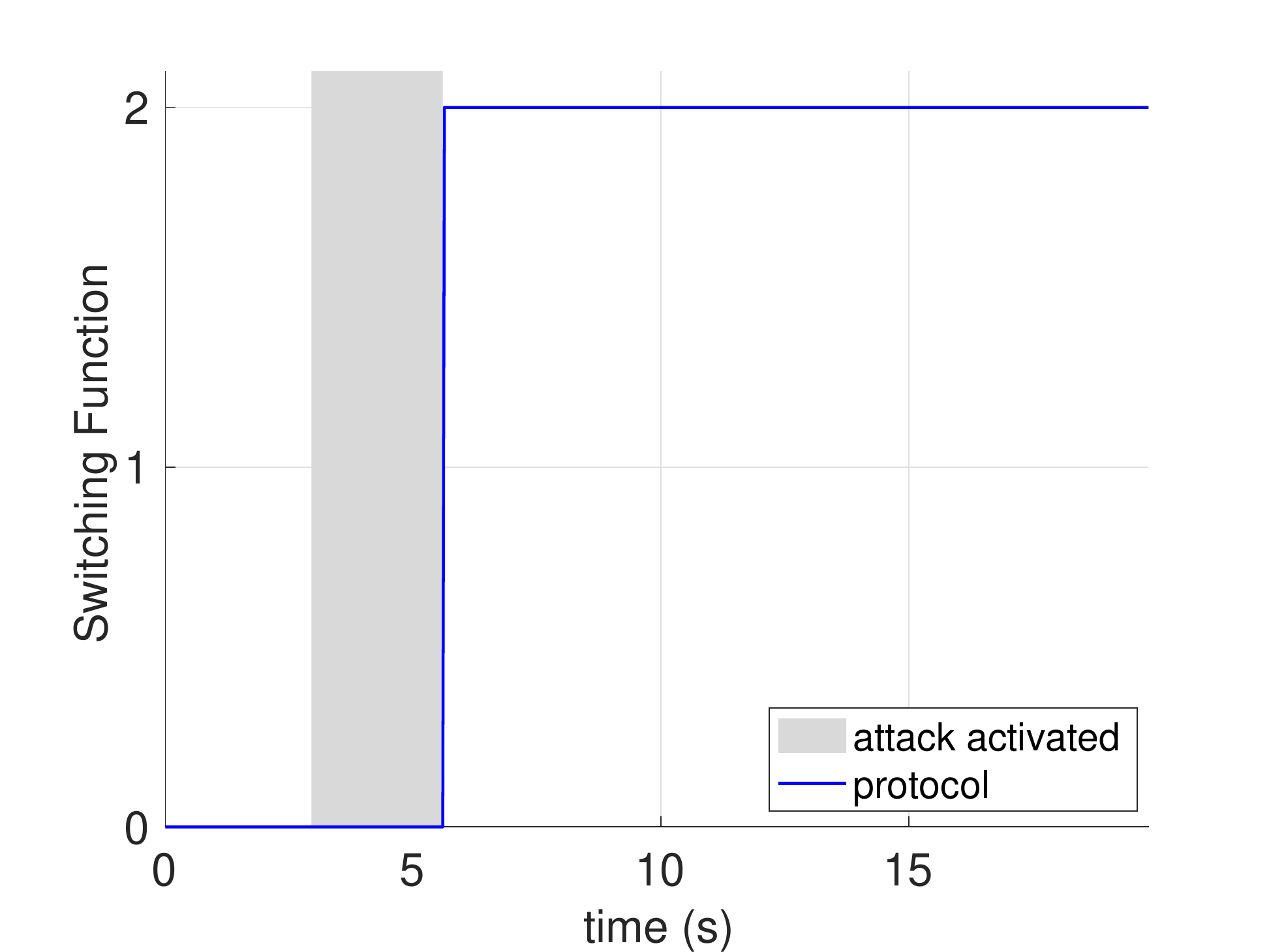}}
    \subfigure[]{\includegraphics[width=0.29\textwidth]{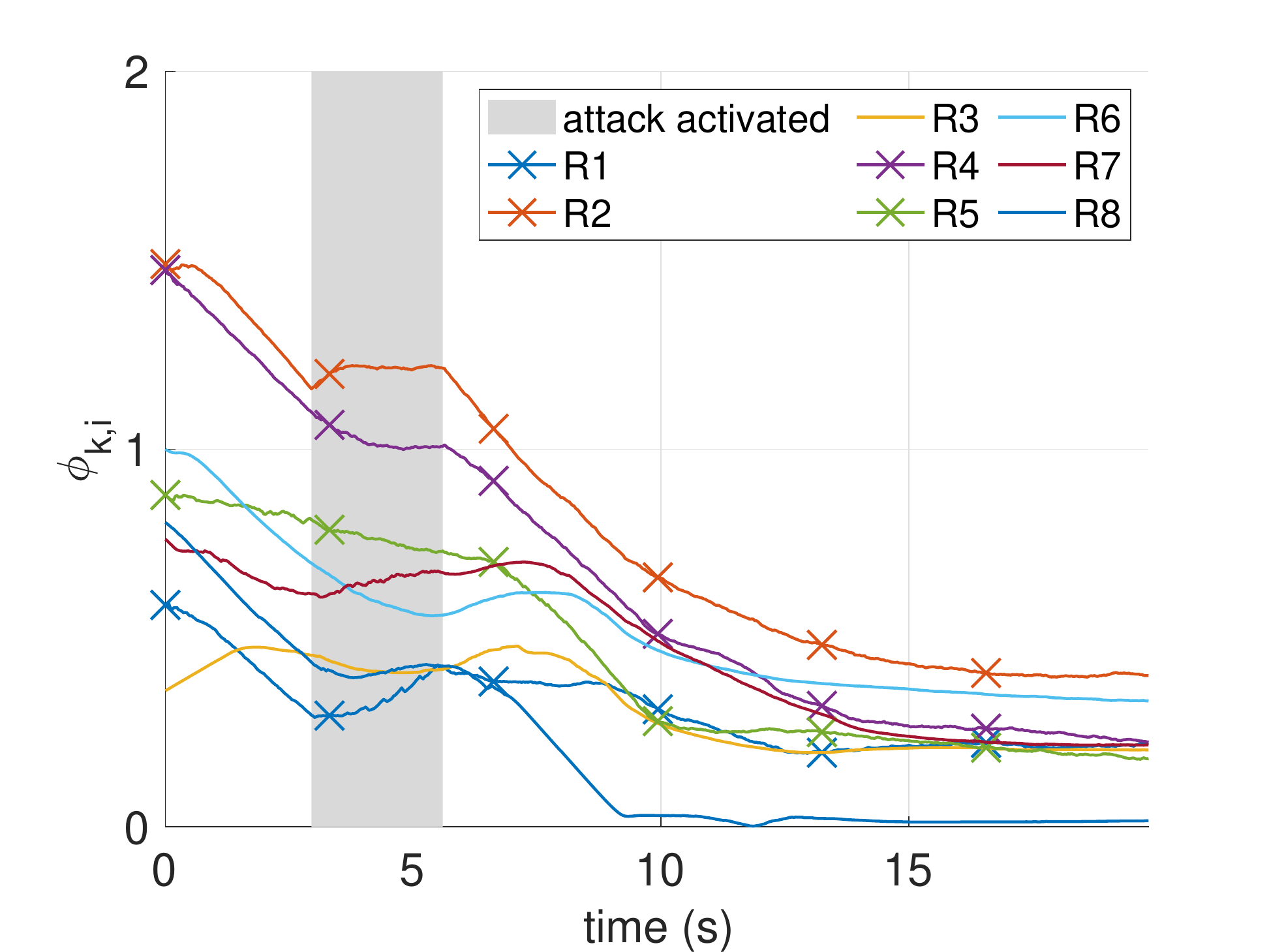}}
    \vspace{-1ex}
	\caption{Multi-robot consensus control in the presence of \textit{denial of service attack} (Robots $1, 2, 4$, and $5$ compromised).}
	\label{fig:C2}
\end{figure*}
\begin{figure*}[t]
	\centering
    \subfigure[]{\includegraphics[width=0.29\textwidth]{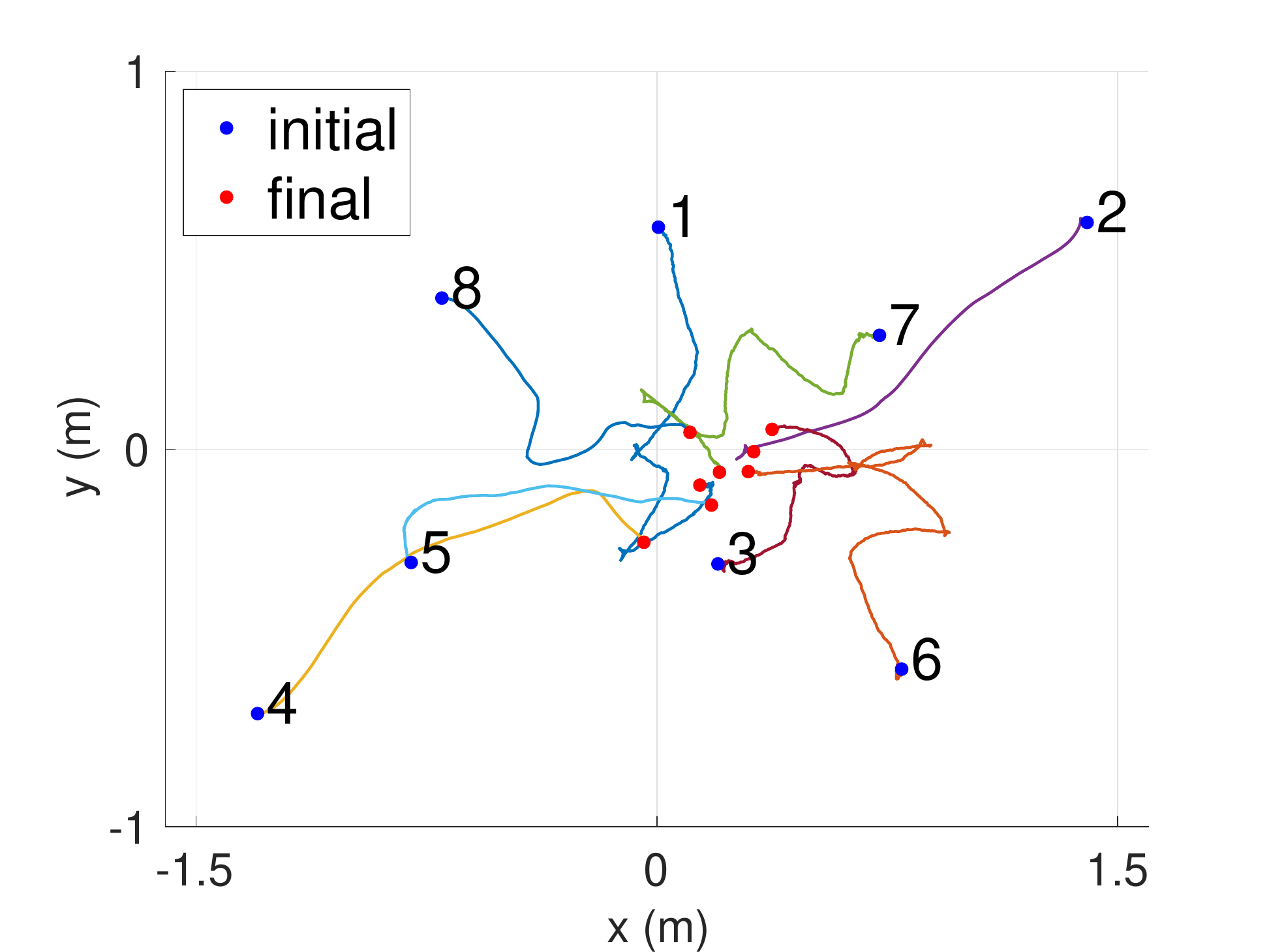}}
    \subfigure[]{\includegraphics[width=0.29\textwidth]{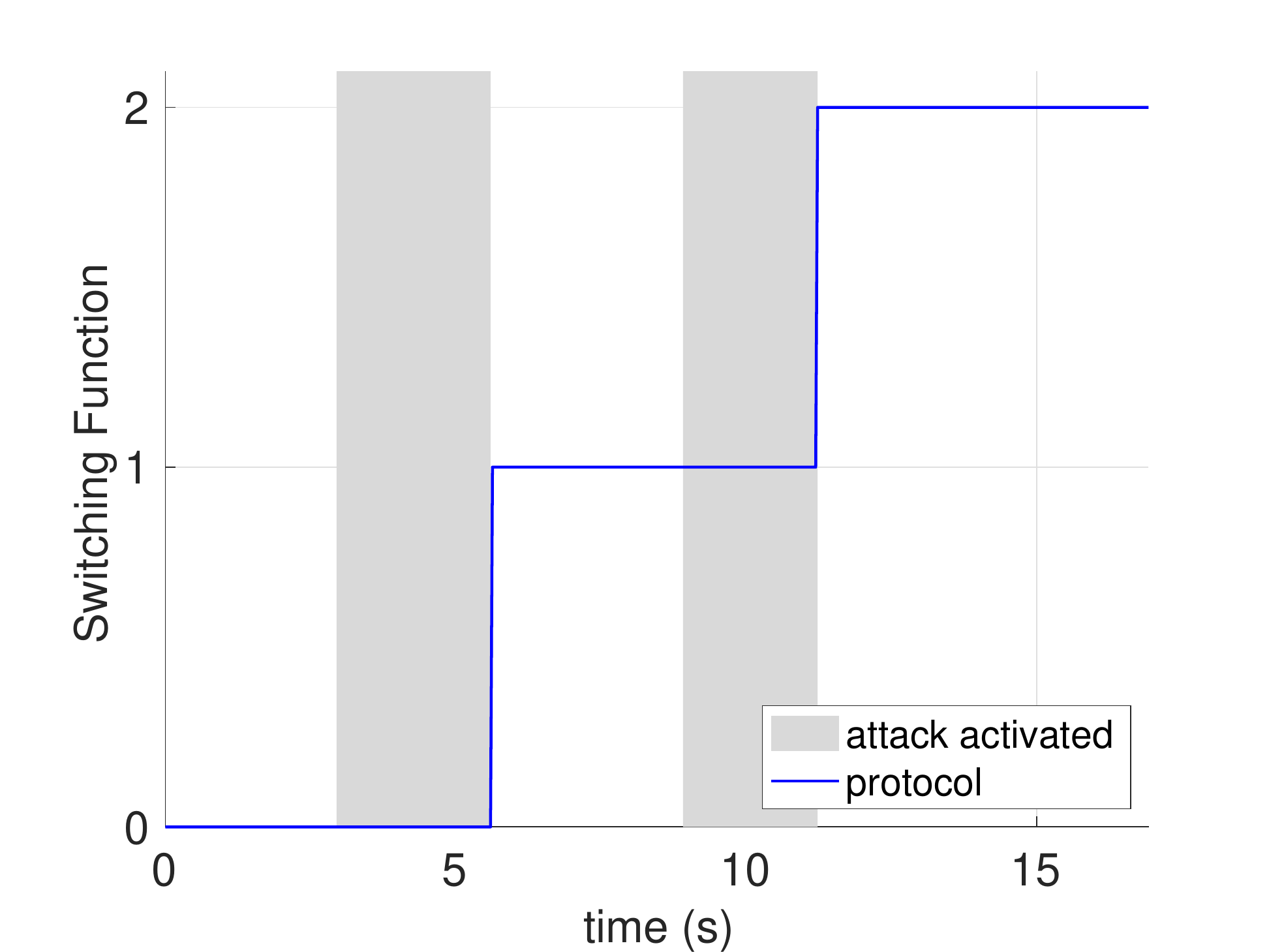}}
    \subfigure[]{\includegraphics[width=0.29\textwidth]{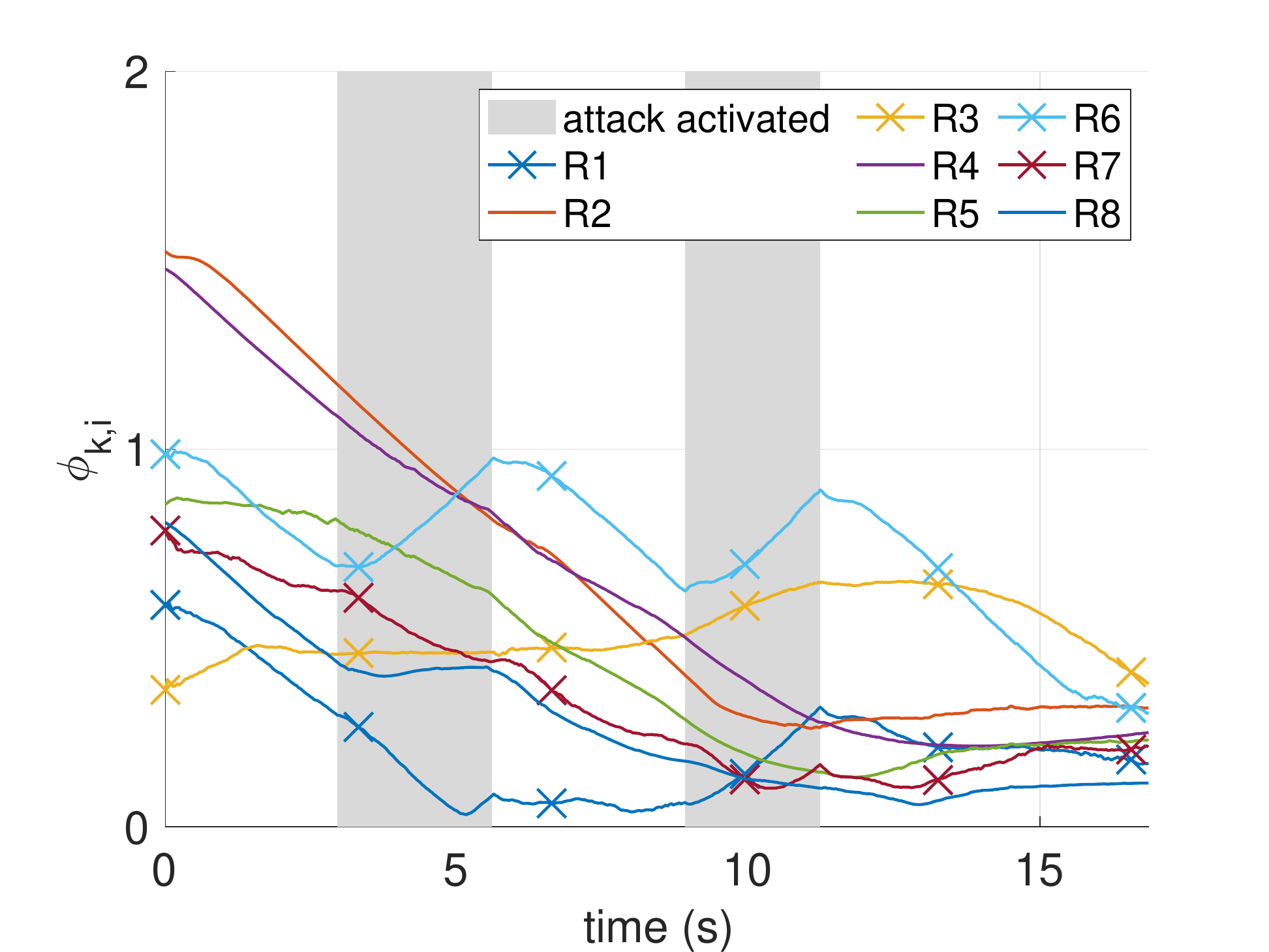}}
    \vspace{-1ex}
	\caption{Multi-robot consensus control in the presence of \textit{deception attack} and \textit{denial of service attack} (Robots $1, 3, 6$, and $7$ compromised).}
    \vspace{-1ex}
	\label{fig:C3}
\end{figure*}

\subsection{Experimental Setup}
For each scenario, eight robots are initialized at a random pose in the $1.6$ m $\times$ $1.0$ m Robotarium testbed. Technical specifications and capabilities of the testbed can be found in \cite{pickem2017robotarium}. Each robot has at least one other robot within its omnidirectional sensor range, $0.8$ m in radius. This condition guarantees that each robot is able to measure the signal strength of other robots at the initial pose. While the team is trying to achieve position consensus at a common point, solitary attacks are injected into four arbitrarily selected robots when the global clock reached $3$ seconds. There exists an upper and lower bound in the magnitude of attacks. Note that the baseline consensus protocol is naturally robust to an attack that is smaller than the detection threshold. A built-in collision avoidance algorithm is automatically executed if $|| x_{i}-x_{j}||<0.1$ m by safe operation requirements (i.e. robots are not allowed to move closer than $0.1$ m). The team is said to achieve global consensus if $||x_{k,i}-x_{k,j}||\leq 0.4$ m as $k \rightarrow \infty$ for all $i,j=1,2,\ldots,N$. The state error covariance $P_{k}$, process noise covariance $Q_{k}$, and measurement noise covariance $R_{k}$ are chosen as the symmetric matrices with the diagonal entries $(10^{-2},10^{-2},10^{-2})$, $(10^{-4},10^{-4},10^{-4})$, and $(10^{-2},10^{-2},10^{-2})$, respectively. The following performance function is introduced to evaluate the overall estimation and control performance for the system (\ref{eq:normal_sys}) and (\ref{eq:ss_attack}):
\begin{equation*}
    \phi_{k,i} = \frac{1}{N} \sum_{k=1}^{N} \|\hat{x}_{k,i}-x_{k,i}\| + \|x_{k,i}^{d}-x_{k,i}\|,
\end{equation*}
where $\hat{x}_{k,i}$ is the estimate of $x_{k,i}$ at time $k$ and $x_{k,i}^{d}$ is the desired state of $i$-th robot achieving consensus.

\subsection{Results}
\begin{figure*}[t]
	\centering
    \subfigure[no attack]{\includegraphics[width=0.245\textwidth]{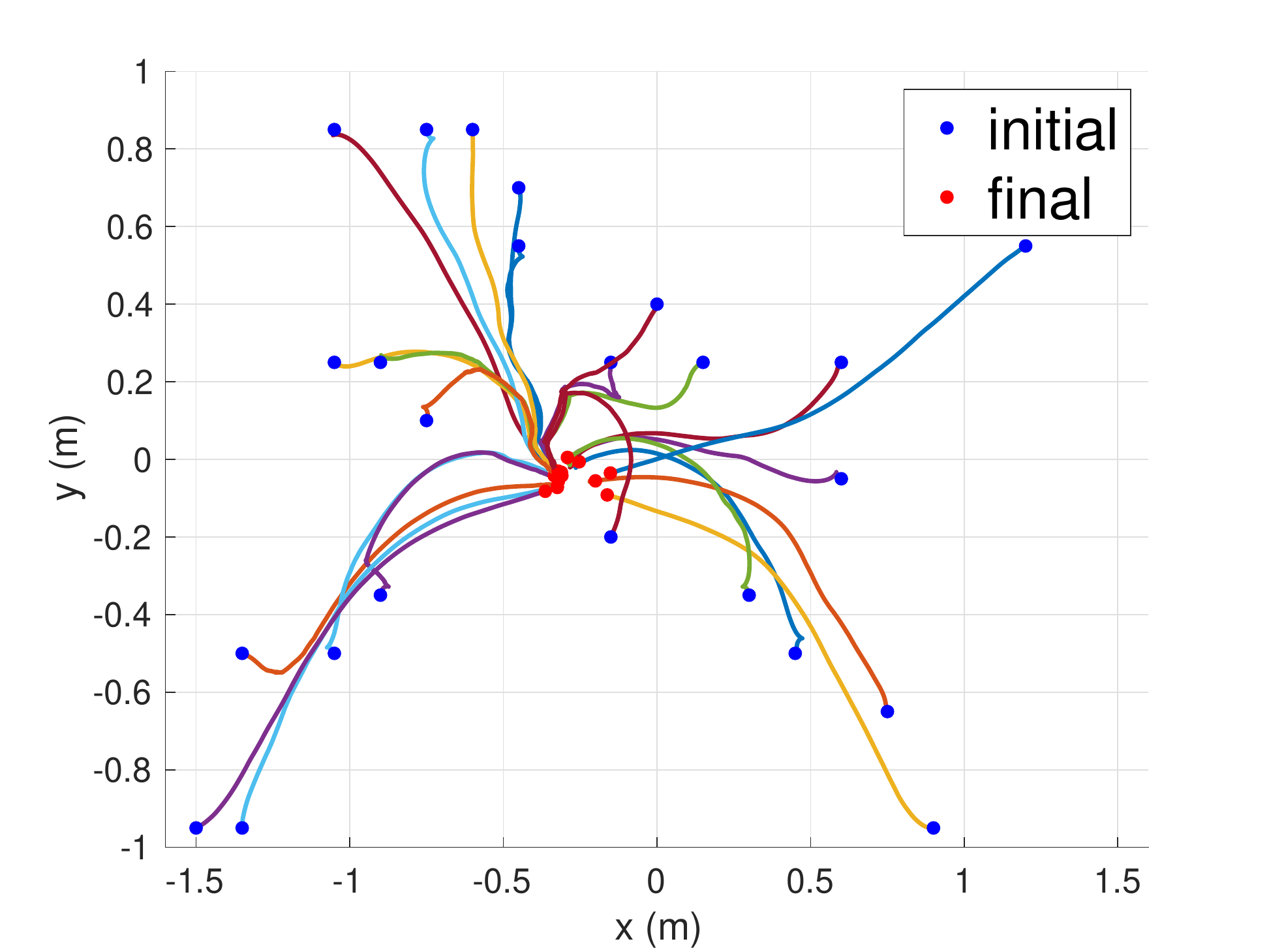}}
    \subfigure[Deception attack]{\includegraphics[width=0.245\textwidth]{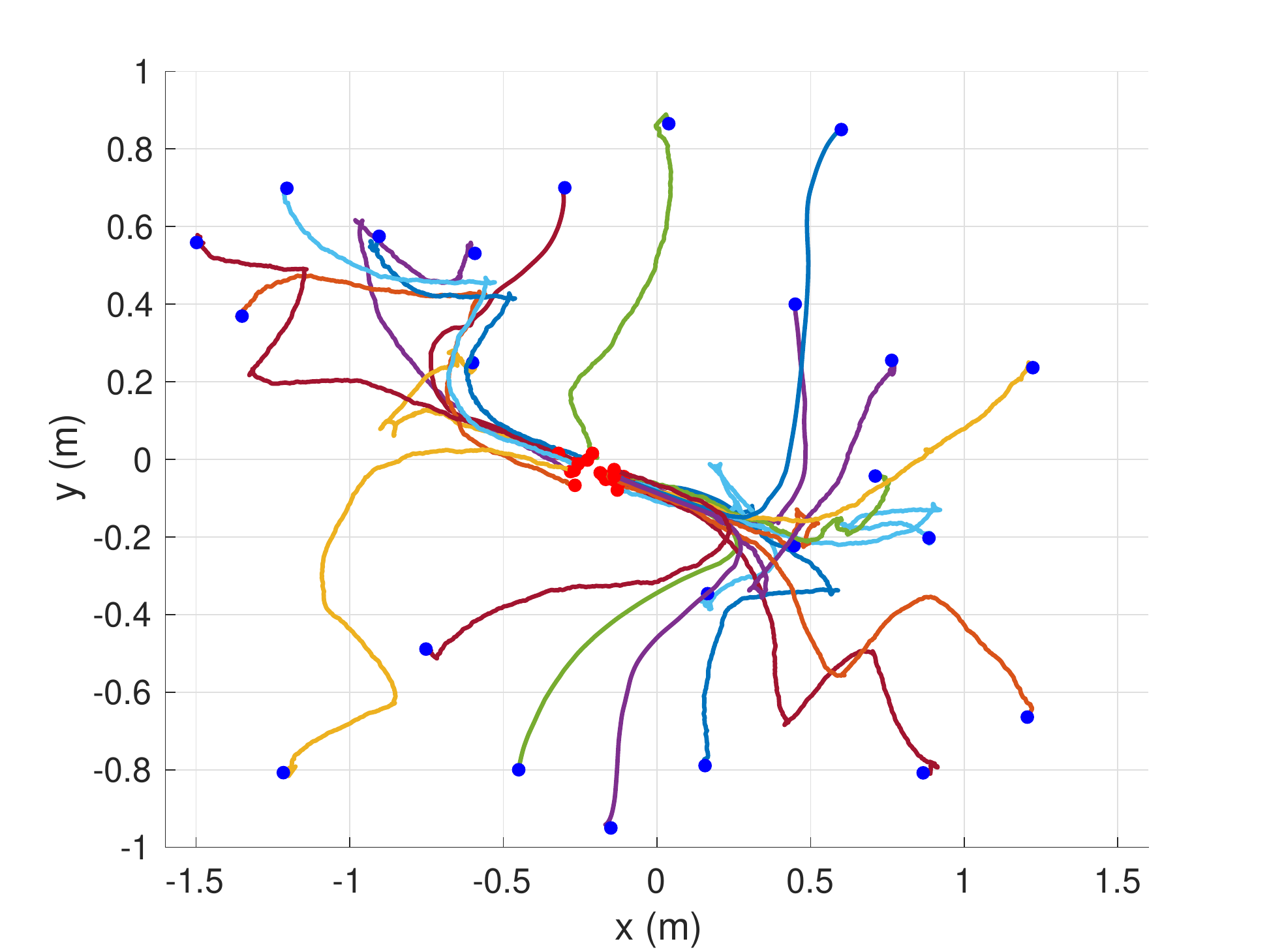}}
    \subfigure[DoS attack]{\includegraphics[width=0.245\textwidth]{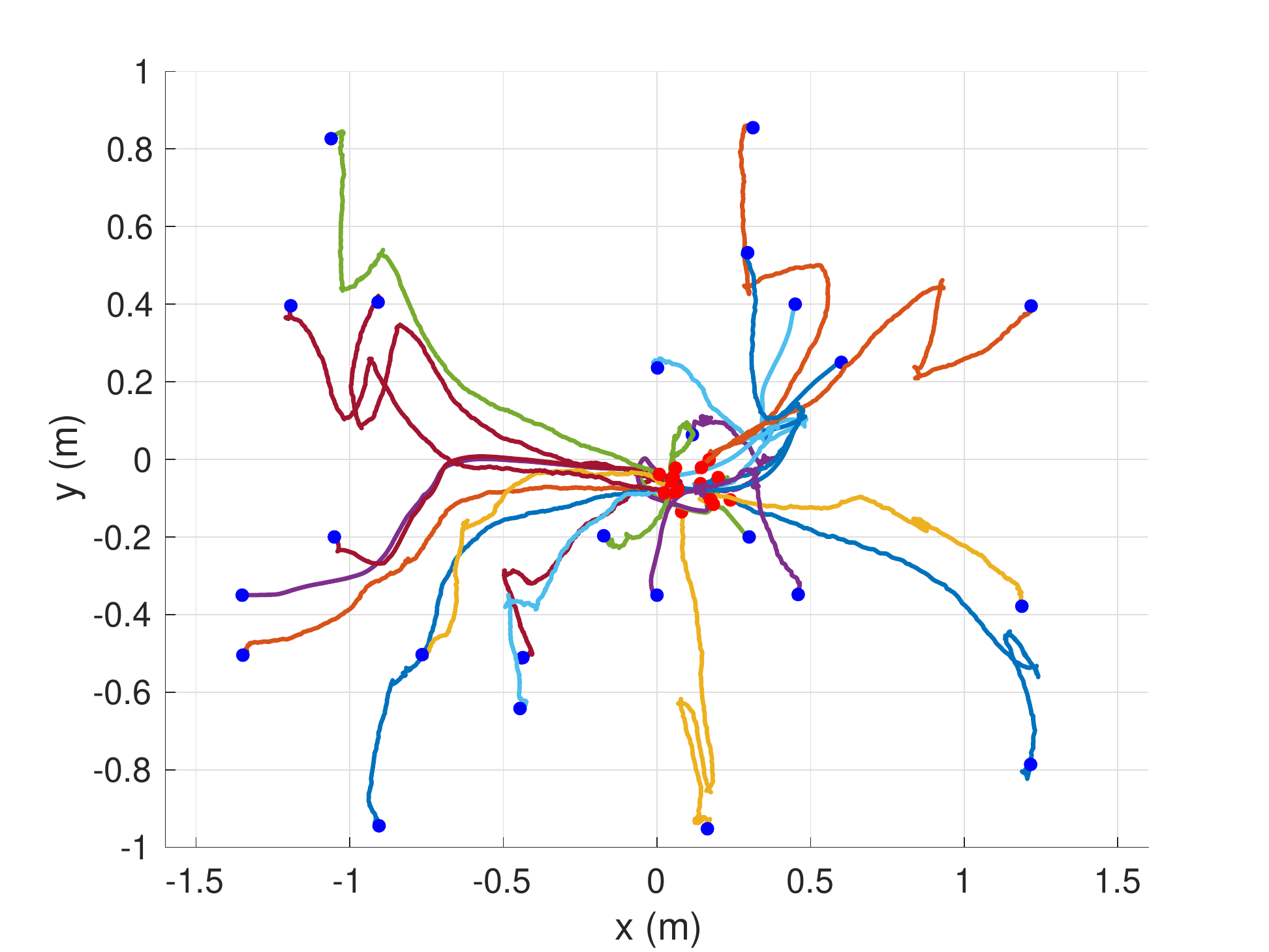}}
    \subfigure[Deception attack]{\includegraphics[width=0.245\textwidth]{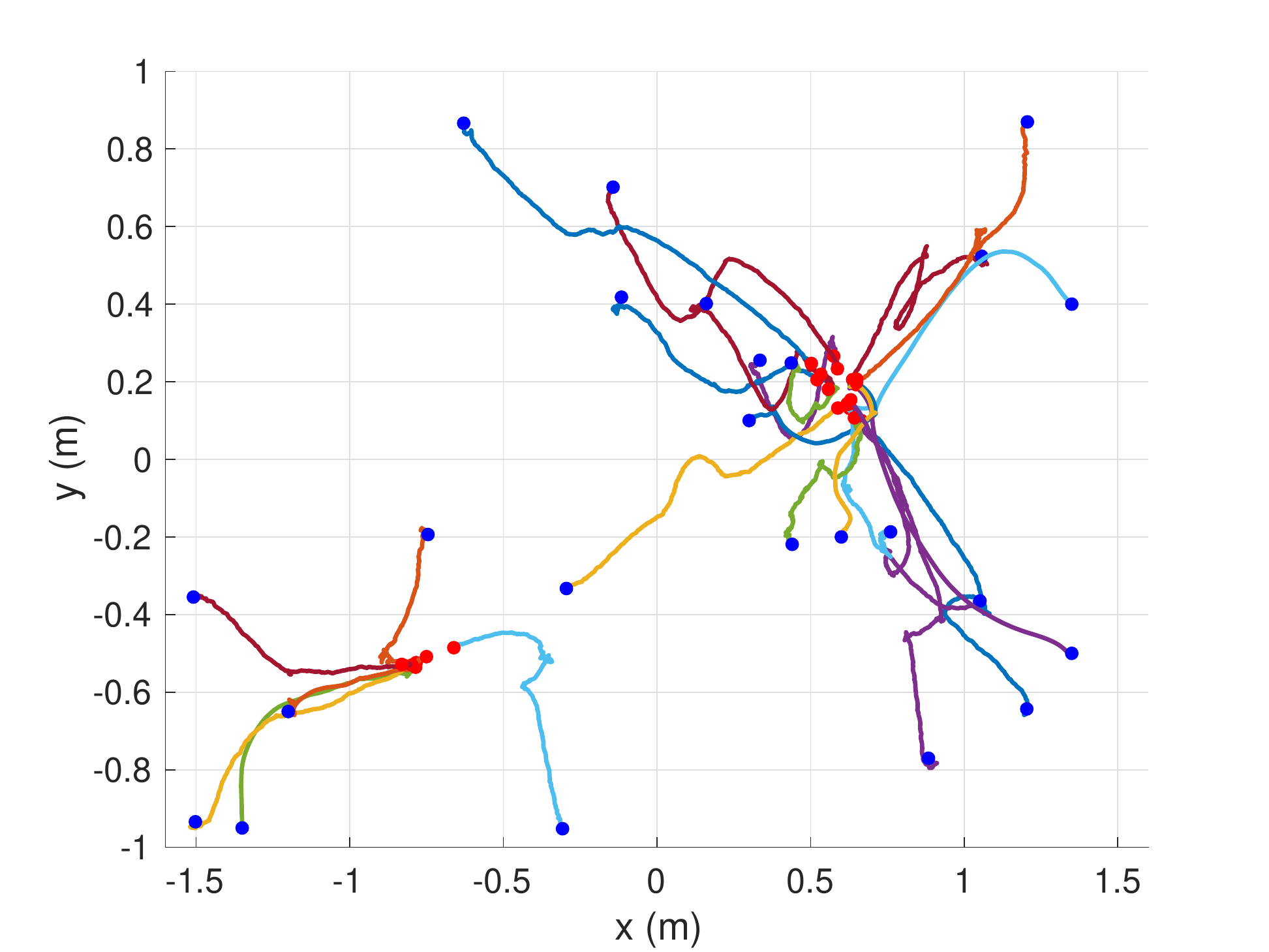}}
	\caption{Multi-robot consensus control in the presence of  18 compromised robots.}
	\label{fig:C3N24}
\end{figure*}
First, an unauthorized user who is capable of compromising the integrity of sensor measurements was considered. This attack rendered an unstable mode of the unobservable system, degrading the performance during the attack period as shown in gray area in Fig. \ref{fig:C1} (c). The performance degradation also can be seen in Fig. \ref{fig:C1} (a) as the compromised robots suddenly deviate from the desired trajectory.
However, the proposed method activated the protocol 1 for countermeasure as illustrated in Fig. \ref{fig:C1} (b) and enabled each robot to verify if the system was functioning properly. It successfully identified attacks on four robots around the $5$ seconds mark when each test statistic went above the threshold. The detection of attacks led the local controller to follow the proposed consensus protocol as soon as the detection scheme confirmed the attacks. In Fig. \ref{fig:C1} (c), the robots eventually achieved position consensus in the presence of more than one compromised robot as the performance function for each robot went less than $0.4$ m. Thus, using the proposed attack-resilient method provided a solution to detect and counteract deception attacks on multiple robots.

The second scenario is that an unauthorized user who is able to compromise the availability of resources. This attack introduced a delay to the communication link between the controllers and actuators. This is observed in Fig. \ref{fig:C2} (c) in which the performance of the compromised robots remained stationary instead of significant change. This is a typical negative effect of delay attack. In Fig. \ref{fig:C2} (b), similar to \textit{Type 1} attack, the detection scheme identified the delay and prompted the local controller to follow the leader-follower consensus protocol. The chronological sequence of robot trajectory is presented in Fig. \ref{fig:C2} (a). As a result, the team reached consensus at a common point in the presence of more than one robot under DoS attacks.

We consider an additional attack scenario that a series of attacks occur in the same experimental setup. A set of \textit{Type 1} attack is first introduced, and then a set of \textit{Type 2} attack is introduced a few seconds later. In Fig. \ref{fig:C3} (b) and (c), the performance degradation can be seen during the attack periods and the switching function activated the corresponding consensus protocol to recover the performance between the attack periods. The global position consensus was achieved in Fig. \ref{fig:C3} (a) as the performance function satisfied the consensus condition in Fig. \ref{fig:C3} (c). In this scenario, more time steps than the previous scenarios were required to achieve consensus as the detection and recovery time increased.

\subsection{Extensive Simulations}
The proposed switching control scheme was applied to more extreme scenarios with different parameters in simulations in order to evaluate its robustness. This was mainly because the maximum number of robots in experiment was limited to $20$, and users were not allowed to remove the default collision avoidance algorithm. 
A hybrid attack was considered where any type of attacks could occur at $\{25,50,75\}$ percentage of total number of robots compromised for $N=8, 12, 16, 20, 24$. Each robot had arbitrary initial conditions and the simulation continued until all robots achieved global state consensus, or the maximum iteration was reached. This set ran for $1,000$ times, and statistical characteristics of the rate to reach state consensus were obtained. 

The statistics show the consensus rate is not significantly affected by attack time or number of compromised robots. However, more failure cases are observed in deception attack type than DoS attack. For example, in Fig. \ref{fig:C3N24}, under random initial conditions for the given $N$, each scenario achieved consensus but partially achieved in Fig. \ref{fig:C3N24} (d). This is because the attack broke the communication link between two groups by deviating from the communication range. While the DoS attack introduced a delay in communication link, resulting the  compromised robots repeated the latest command (in this case, the robots kept moving toward where they headed to), the deception attack introduced a set of false data, making the compromised robots to believe the false data as a new command until it was identified as an attack (in this case, the robots significantly deviated from the desired path).
Though we only present the simulation results where 18 of 24 robots compromised due to the space limitation, the switching consensus control scheme achieved state consensus for all cases in which the detection mechanism completed the attack identification. A full demonstration is available in the supplementary video.

\section{Conclusion}\label{sec:conclusion}
In this paper, we presented a switching control scheme that ensures multi-robot consensus problem in the presence of multiple compromised robots. Our key approach is using a weighted bearing controller and a leader-follower strategy to cope with two major types of cyberattacks. From the extensive experiments, we demonstrated that the proposed approach allows multi-robots to achieve consensus. In the simulations, the proposed method achieved consensus even for the cases where the majority of robots in a team are attacked (75$\%$). In the near future, this study will be extended to deal with more sophisticated scenarios, such as different types of attacks at the same time and disguised attacks that do not significantly affect the residual but still affect the system.

\section*{Acknowledgment}
This work was supported by Award No. 2017-R2-CX-0001, awarded by the National Institute of Justice, Office of Justice Programs, U.S. Department of Justice. The opinions, findings, and conclusions or recommendations expressed in this paper are the authors’ and do not necessarily reflect those of the Department of Justice.

\bibliographystyle{IEEEtran}
\bibliography{references}

\end{document}